\documentclass[letterpaper, 10 pt, conference]{ieeeconf}
\IEEEoverridecommandlockouts
\usepackage{amsmath,amssymb,amsfonts,mathtools}
\usepackage{algorithmic}
\usepackage{algorithm}
\usepackage{graphicx}
\usepackage{textcomp}
\usepackage{booktabs}
\usepackage{cite}
\usepackage{url}

\begin{document}

\title{\LARGE \bf Auction-Based Task Allocation with Energy-Conscientious Trajectory Optimization for AMR Fleets}

\author{
Jiachen Li$^{1}$, Soovadeep Bakshi$^{1}$, Jian Chu$^{1}$, Shihao Li$^{1}$, and Dongmei Chen$^{1}$%
\thanks{$^{1}$All authors are with the Department of Mechanical Engineering, The University of Texas at Austin, Austin, TX 78712, USA.
{\tt\small \{jiachenli, soovadeepbakshi, jian\_chu, shihaoli01301\}@utexas.edu, dmchen@me.utexas.edu}}%
}

\maketitle

\begin{abstract}
This paper presents a hierarchical two-stage framework for multi-robot task allocation and trajectory optimization in asymmetric task spaces: (1)~a sequential auction allocates tasks using closed-form bid functions, and (2)~each robot independently solves an optimal control problem for energy-minimal trajectories with a physics-based battery model, followed by a collision avoidance refinement step using pairwise proximity penalties. Event-triggered warm-start rescheduling with bounded trigger frequency handles robot faults, priority arrivals, and energy deviations. Across 505 scenarios with 2–20 robots and up to 100 tasks on three factory layouts, both energy- and distance-based auction variants achieve 11.8\% average energy savings over nearest-task allocation, with rescheduling latency under 10 ms. The central finding is that bid-metric performance is regime-dependent: in uniform workspaces, distance bids outperform energy bids by 3.5\% ($p < 0.05$, Wilcoxon) because a 15.7\% closed-form approximation error degrades bid ranking accuracy to 87\%; however, when workspace friction heterogeneity is sufficient ($r < 0.85$ energy-distance correlation), a zone-aware energy bid outperforms distance bids by 2–2.4\%. These results provide practitioner guidance: use distance bids in near-uniform terrain and energy-aware bids when friction variation is significant.
\end{abstract}

\section{Introduction}

Autonomous mobile robots (AMRs) are increasingly deployed in manufacturing and warehouse environments for material transport. As fleet sizes grow, minimizing total energy consumption becomes critical for reducing operational costs, extending battery life, and decreasing charging downtime~\cite{bakshi2019fast,mei2006energy}. In multi-robot fleets, total energy is determined jointly by two coupled decisions: \emph{which} robot executes each task (allocation) and \emph{how} each robot traverses its route (trajectory optimization). The current approach is to handle these two aspects separately: first, tasks are assigned based on distance or time, and then a path is planned for each robot. This approach overlooks the interaction between the quality of task assignment and energy consumption at the path planning level.

The multi-robot task allocation (MRTA) problem assigns tasks to robots to optimize a fleet-level objective~\cite{korsah2013taxonomy}. Auction-based approaches, including sequential single-item auctions~\cite{koenig2006power} and distributed algorithms~\cite{luo2015distributed}, offer tractable solutions with approximation guaranties, yet they almost universally adopt Euclidean distance or completion time as the bid metric. Nunes et al.~\cite{nunes2017taxonomy} comprehensively survey allocation strategies and note the absence of physics-based cost models in auction mechanisms. This reliance on geometric proxies assumes that energy cost is monotonically related to distance—an assumption that breaks down in \emph{asymmetric task spaces}, where traversal cost depends on payload mass, travel direction, velocity boundary conditions, and spatially varying terrain friction.

Physics-based energy modeling for mobile robots has received growing attention~\cite{mei2006energy,behrens2023energy}. Bakshi et al.~\cite{bakshi2021energy} formulated an optimal control problem (OCP) for a single AMR that minimizes a cost functional accounting for rolling friction, acceleration dynamics, motor efficiency, and battery electrochemistry in asymmetric task spaces. At the fleet level, energy-aware scheduling has been addressed through meta-heuristic optimization~\cite{behrens2023energy} and consensus protocols~\cite{nunes2017taxonomy}, while dynamic rescheduling has been tackled via multi-agent reinforcement learning~\cite{huttenrauch2019deep}, event-driven evolutionary methods~\cite{segui2024event}, and the FLEET framework~\cite{chen2025fleet}. However, these fleet-level approaches rely on simplified energy surrogates—typically distance-proportional or battery-level approximations—that discard the direction-, payload-, and velocity-dependent cost structure of physics-based models.

A fundamental question therefore remains: \emph{does embedding a physics-based energy model into the allocation mechanism yield measurable fleet-level savings, or does the surrogate approximation error negate the richer cost information?} No prior work simultaneously integrates (a)~physics-based AMR energy models as auction bid functions with (b)~asymmetric, direction-dependent cost preservation and (c)~event-triggered warm-start rescheduling with bounded trigger frequency. Nor does any existing study provide a controlled comparison of energy-based versus distance-based bids within a single framework, identifying when each dominates.

This paper addresses these gaps with a \emph{hierarchical two-stage framework}. A sequential auction first allocates tasks using closed-form energy bid functions; each robot then solves the full nonlinear OCP for its assigned sequence, followed by pairwise collision avoidance refinement. An event-triggered rescheduling layer handles robot faults, priority arrivals, and energy deviations. Across 505 scenarios on three factory layouts with 2–20 robots and up to 100 tasks, we characterize when energy bids outperform distance metrics.

\begin{figure}[t]
\centering
\includegraphics[width=\columnwidth]{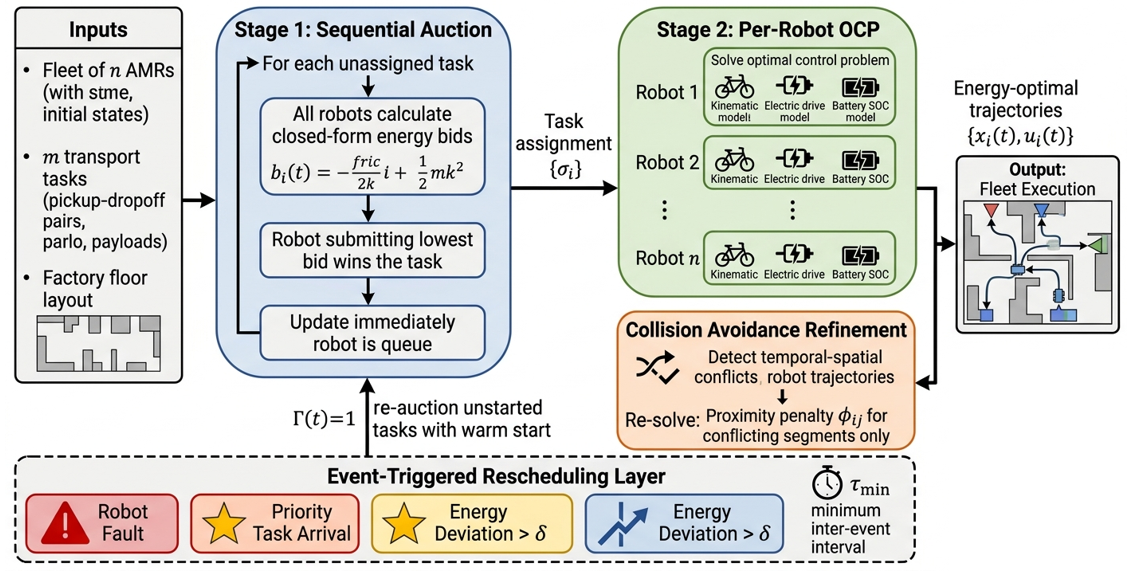}
\caption{Overview of the proposed two-stage auction-plus-OCP framework with collision avoidance refinement and event-triggered rescheduling.}
\label{fig:overview}
\end{figure}

The contributions are: (i)~a hierarchical framework integrating the physics-based formulation of~\cite{bakshi2021energy} into fleet-level allocation with post-allocation collision avoidance via proximity penalties; (ii)~11.8\% fleet energy savings over nearest-task allocation with event-triggered rescheduling under 10\,ms latency; and (iii)~an empirical characterization of the bid-metric crossover, identifying friction heterogeneity ($r \approx 0.85$ energy--distance correlation) as the decisive variable: distance bids dominate in uniform terrain due to 15.7\% approximation error degrading ranking accuracy, while zone-aware energy bids recover 2--2.4\% savings when friction variation is sufficient. Fig.~\ref{fig:overview} provides a schematic overview.

\section{Problem Formulation}

\subsection{Notation}

Consider a fleet of $n$~AMRs, $\mathcal{N} = \{1, \ldots, n\}$, assigned to execute $m$~transport tasks, $\mathcal{T} = \{1, \ldots, m\}$. Each task $t \in \mathcal{T}$ is an ordered pair $(p_t, d_t)$ where $p_t$ is the pickup location and $d_t$ is the dropoff location. Let $\sigma_i = (t_{i,1}, t_{i,2}, \ldots, t_{i,k_i})$ denote the ordered task sequence assigned to robot~$i$, and let $E_i(\sigma_i)$ denote the total energy consumed by robot~$i$ while executing that sequence. The workspace is asymmetric because the transition cost from one task to the next depends jointly on unloaded versus loaded motion, travel direction, and velocity boundary conditions inherited from the previous and next phases.

For robot~$i$, the state is
\begin{equation}
x_i = [X_i,\; Y_i,\; \psi_i,\; v_{x,i},\; \mathrm{SOC}_i]^\top \in \mathbb{R}^5
\label{eq:statei}
\end{equation}
and the control is
\begin{equation}
u_i = [\delta_{f,i},\; V_{m,i},\; \tau_{b,i}]^\top \in \mathbb{R}^3.
\label{eq:controli}
\end{equation}

\subsection{Inherited Single-Robot Optimal Control Formulation}

The revised formulation preserves the single-robot mathematical structure and uses it as the continuous layer of the fleet framework. The task schedule for each AMR is assumed to be known within the single-robot subproblem; the kinematic model is taken to be sufficiently accurate in the operating regime, and Dubins geometry is used to define the nominal segment timing between boundary waypoints.

\begin{figure}[t]
\centering
\includegraphics[width=0.75\columnwidth]{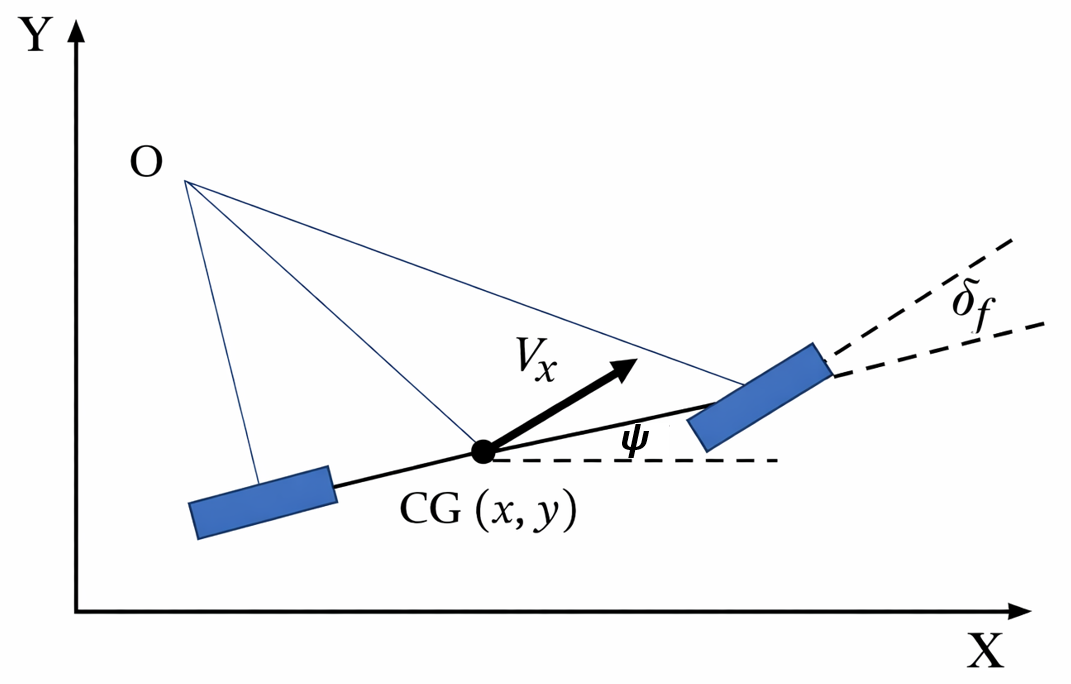}
\caption{Bicycle Model for AMR.}
\label{fig:bicycle}
\end{figure}

The simplified kinematic bicycle model (Fig.~\ref{fig:bicycle}) is
\begin{align}
\dot X &= v_x \cos(\psi) \label{eq:sr_dyn1}\\
\dot Y &= v_x \sin(\psi) \label{eq:sr_dyn2}\\
\dot \psi &= \frac{v_x}{L}\tan(\delta_f) \label{eq:sr_dyn3}\\
\dot v_x &= \frac{F_x}{m}. \label{eq:sr_dyn4}
\end{align}

The electric drive model is written as
\begin{align}
i &= \frac{V_m - K_m \frac{v_x}{r_w}}{R_m} \label{eq:sr_act1}\\
\tau_m &= K_m i \label{eq:sr_act2}\\
\dot v_x &= \frac{\tau_m - \tau_b}{m r_w \left(1 + \frac{J_m}{m r_w^2}\right)}. \label{eq:sr_act3}
\end{align}
The battery model is
\begin{align}
\mathrm{OCV} &= a_1 e^{b_1 \mathrm{SOC}} + a_2 e^{b_2 \mathrm{SOC}} + c\,\mathrm{SOC}^2 \label{eq:sr_batt1}\\
P_{\mathrm{battery}} &= \mathrm{OCV}\cdot i_{\mathrm{battery}} = P_{\mathrm{demand}} + P_{\mathrm{loss}} \label{eq:sr_batt2}\\
\dot{\mathrm{SOC}} &= -\frac{i_{\mathrm{battery}}}{Q} \label{eq:sr_batt3}\\
P_{\mathrm{battery}} &= \frac{P_{\mathrm{demand}}}{\eta\!\left(1+e^{-P_{\mathrm{demand}}}\right)} + \frac{P_{\mathrm{demand}}\eta}{1+e^{P_{\mathrm{demand}}}}. \label{eq:sr_batt4}
\end{align}
Equations~\eqref{eq:sr_dyn1}--\eqref{eq:sr_batt4} define the nonlinear physics model used throughout the paper.

Let the single-robot state be $x = [X,\;Y,\;\psi,\;v_x,\;\mathrm{SOC}]^\top$ and control be $u = [\delta_f,\;V_m,\;\tau_b]^\top$. The inherited energy-aware single-robot objective is
\begin{equation}
J^{(1)}(x,u) = \int_0^{t_f} \left(w_1 P_{\mathrm{batt}} + w_2(\mathrm{SOC}_{\max}-\mathrm{SOC})^2 + w_3 \dot{\psi}^{\,2}\right) dt.
\label{eq:single_ocp_cost}
\end{equation}
The corresponding multiphase single-robot optimal control problem is
\begin{align}
\min_{x(\cdot),u(\cdot),\{t_\ell\}} \quad & J^{(1)}(x,u) \label{eq:single_ocp_main}\\
\textrm{s.t.}\quad & \dot x = f(x,u), \label{eq:single_ocp_dyn}\\
& X_{\min}\le X(t)\le X_{\max}, \notag\\
& Y_{\min}\le Y(t)\le Y_{\max}, \label{eq:single_ocp_box}\\
& h_{\min}\le h(X(t),Y(t))\le h_{\max}, \label{eq:single_ocp_height}\\
& 0\le v_x(t)\le v_{\max}, \notag\\
& \mathrm{SOC}_{\min}\le \mathrm{SOC}(t)\le \mathrm{SOC}_{\max}, \label{eq:single_ocp_state}\\
& 0\le V_m(t)\le V_{m,\max}, \notag\\
& 0\le \tau_b(t)\le \tau_{b,\max}, \label{eq:single_ocp_ctrl1}\\
& -\delta_{f,\max}\le \delta_f(t)\le \delta_{f,\max}, \label{eq:single_ocp_ctrl2}\\
& x(0) = [X_d,\,Y_d,\,0,\,0,\,\mathrm{SOC}_{\max}]^\top, \label{eq:single_ocp_bc1}\\
& x(t_\ell)= [X_\ell,\,Y_\ell,\,\psi_\ell,\,0,\,\mathrm{SOC}_\ell]^\top, \notag\\
& \hspace{8em} \ell=1,\dots,2n_t, \label{eq:single_ocp_bc2}\\
& x(t_f)=x(t_{2n_t+1}) \notag\\
& \quad = [X_d,\,Y_d,\,0,\,0,\,\mathrm{SOC}_f]^\top, \label{eq:single_ocp_bc3}\\
& t_\ell = t_{\ell-1} \notag\\
& \quad + \frac{\mathrm{DL}([X,Y,\psi]_{\ell-1},[X,Y,\psi]_{\ell})}{v_{\mathrm{avg},\ell}}, \notag\\
& \hspace{8em} \ell=1,\dots,2n_t\!+\!1. \label{eq:single_ocp_time}
\end{align}
Here $n_t$ denotes the number of assigned tasks for the single robot, and each waypoint is either a pickup or a dropoff boundary. Equation~\eqref{eq:single_ocp_time} is the inherited timing law that binds the multiphase boundary conditions to the average-speed constraint. The reduced notation used later in the paper is unchanged:
\begin{equation}
\min_{u(t)} \int_0^T P\!\left(x(t), u(t)\right) dt
\label{eq:ocp}
\end{equation}
subject to robot dynamics $\dot{x} = f(x, u)$ and boundary conditions at task pickup and dropoff locations.

\subsection{Formal Multi-Agent Extension}

The fleet-level problem introduces one copy of the inherited single-robot dynamics per robot and couples those copies through assignment, ordering, and coordination constraints. Let
\begin{equation}
\begin{aligned}
&Y = [y_{it}] \in \{0,1\}^{n\times m}, \\
&y_{it}=1 \iff \text{task } t \text{ is assigned to robot } i.
\end{aligned}
\label{eq:assignbin}
\end{equation}
The partition constraints are
\begin{align}
\sum_{i=1}^n y_{it} &= 1, \qquad t=1,\dots,m \label{eq:partition1}\\
y_{it} &\in \{0,1\}, \qquad i=1,\dots,n,\;\; t=1,\dots,m. \label{eq:partition2}
\end{align}
For each robot~$i$, let $\pi_i$ be an ordering of the tasks assigned to that robot, so that $\pi_i$ is a permutation of the set $\{t\in\mathcal{T}: y_{it}=1\}$. The multiphase waypoint list induced by $\pi_i$ is denoted
\begin{equation}
\mathcal{W}_i(\pi_i)=\left(q_i^0,\; p_{\pi_i(1)},\; d_{\pi_i(1)},\;\ldots,\; p_{\pi_i(k_i)},\; d_{\pi_i(k_i)},\; q_i^f\right),
\label{eq:waypoints}
\end{equation}
where $q_i^0$ and $q_i^f$ are the depot states.

The exact fleet-level problem is
\begin{align}
\min_{\substack{Y,\{\pi_i\},\\ \{x_i(\cdot),u_i(\cdot)\}_{i=1}^n}} \; & J_{\mathrm{fleet}}
:= \sum_{i=1}^n J_i\!\bigl(x_i,u_i;\mathcal{W}_i(\pi_i)\bigr) \notag\\
& + \lambda_c \!\!\sum_{1\le i<j\le n} \int_0^{T_f} \!\phi_{ij}(x_i,x_j)\,dt
\label{eq:fleet_problem}\\
\textrm{s.t.}\quad & \dot x_i = f_i(x_i,u_i), \notag\\
& \hspace{8em} i=1,\dots,n, \label{eq:fleet_dyn}\\
& x_i(\cdot),u_i(\cdot) \textrm{ satisfy } \notag\\
& \quad \eqref{eq:single_ocp_box}\textrm{--}\eqref{eq:single_ocp_time}
  \textrm{ on } \mathcal{W}_i(\pi_i), \label{eq:fleet_single_constraints}\\
& \eqref{eq:partition1},\eqref{eq:partition2} \textrm{ hold.} \label{eq:fleet_partition_constraints}
\end{align}
The function $\phi_{ij}$ is a nonnegative proximity penalty that enforces inter-robot collision avoidance. We adopt a soft barrier formulation:
\begin{equation}
\phi_{ij}(x_i,x_j) = \max\!\left(0,\; 1 - \frac{\|[X_i,Y_i]^\top - [X_j,Y_j]^\top\|}{d_{\mathrm{safe}}}\right)^{\!2}
\label{eq:collision}
\end{equation}
where $d_{\mathrm{safe}}$ is the minimum safe separation distance. This penalty is zero whenever robots are farther apart than $d_{\mathrm{safe}}$ and grows quadratically as they approach. In the hierarchical decomposition used in this paper, collision avoidance is enforced in a post-allocation refinement step: after the auction assigns tasks and each robot solves its independent OCP, a pairwise check identifies temporal–spatial conflicts, and conflicting trajectory segments are re-solved with the $\phi_{ij}$ coupling active ($\lambda_c > 0$). This hierarchical approach preserves the $O(nm^2)$ auction scalability while handling physical inter-robot interactions at the trajectory level. Note that allocation decisions do not internalize downstream interaction costs exactly; the collision penalty is resolved only after task assignment, so the framework is a hierarchical decomposition rather than a jointly optimal fleet coordination scheme.

\subsection{Asymmetric Cost Matrix with Payload and Velocity Boundary Conditions}

The asymmetry of the task space is formalized by the transition matrix induced by the inherited single-robot OCP. The per-segment energy for a robot of mass~$M$ carrying payload~$w$ traversing distance~$s$ is
\begin{equation}
\begin{aligned}
E_{\mathrm{seg}}(s,w) = \frac{1}{\eta}\Big[&\mu(M\!+\!w)g\,s \\
&+ \tfrac{1}{2}(M\!+\!w)\bigl(v_f^2 - v_0^2\bigr)\Big]
\end{aligned}
\label{eq:eseg}
\end{equation}
where $\mu$ is the rolling friction coefficient, $g$ is gravitational acceleration, $\eta$ is motor efficiency, and $v_0, v_f$ are initial and final velocities. Whenever boundary-speed dependence must be emphasized, the same quantity is written as $E_{\mathrm{seg}}(s,w;v_0,v_f)$.

When regenerative braking is active, a fraction~$\eta_r$ of kinetic energy is recovered during deceleration, and $E_{\mathrm{seg}}$ can become negative for segments where deceleration dominates. When regenerative braking is disabled, we enforce $E_{\mathrm{seg}} = \max(E_{\mathrm{seg}}^{\mathrm{raw}}, 0)$.

The total per-robot energy is
\begin{align}
E_i(\sigma_i) = \sum_{j=1}^{k_i} \Big[&\; E_{\mathrm{seg}}\!\left(\|d_{t_{i,j-1}} - p_{t_{i,j}}\|,\; 0\right) \nonumber\\
&+ E_{\mathrm{seg}}\!\left(\|p_{t_{i,j}} - d_{t_{i,j}}\|,\; w_{t_{i,j}}\right) \Big]
\label{eq:etotal}
\end{align}
where the first term is \emph{unloaded} transit ($w = 0$) from the previous dropoff to the next pickup, and the second is \emph{loaded} task execution with payload $w = w_{t_{i,j}}$. This asymmetric structure---different payload masses and different distances for the two legs of each task---captures why the total energy of a route is not invariant under reversal.

To make the dependence on payload and velocity boundary conditions explicit at the OCP level, define the ordered-pair transition cost for robot~$i$ as
\begin{equation}
\begin{aligned}
&C_i(a,b;w,v^-,v^+) := \min_{x_i,u_i,T\ge 0} \int_0^{T} \ell_i \, dt, \\
&\ell_i = w_1 P_{\mathrm{batt},i} + w_2(\mathrm{SOC}_{i,\max}-\mathrm{SOC}_i)^2 + w_3 \dot{\psi}_i^{\,2}
\end{aligned}
\label{eq:Ci}
\end{equation}
subject to \eqref{eq:single_ocp_dyn}--\eqref{eq:single_ocp_ctrl2} with
\begin{equation}
\begin{aligned}
[X_i(0),Y_i(0)]^\top &= a, \quad v_{x,i}(0)=v^-, \\
[X_i(T),Y_i(T)]^\top &= b, \quad v_{x,i}(T)=v^+,
\end{aligned}
\label{eq:Ci_bc}
\end{equation}
and payload parameter $w$ embedded in the demand power and mass terms. The asymmetric transition matrix between tasks $t_a$ and $t_b$ is then
\begin{equation}
\mathcal{A}_i(t_a,t_b)
:=
C_i\!\left(d_{t_a},p_{t_b};0,v_{a}^{+},v_{b}^{-}\right)
+
C_i\!\left(p_{t_b},d_{t_b};w_{t_b},v_{b}^{-},v_{b}^{+}\right).
\label{eq:asym_matrix}
\end{equation}
Because the payload argument and boundary velocities change with direction and task identity, one generally has
\begin{equation}
\mathcal{A}_i(t_a,t_b)\neq \mathcal{A}_i(t_b,t_a).
\label{eq:asymmetry}
\end{equation}
Equation~\eqref{eq:asym_matrix} is the fleet-level asymmetric cost matrix used conceptually by the assignment layer. The sequential auction replaces the expensive exact cost~\eqref{eq:asym_matrix} with the tractable approximation introduced next.

\subsection{Closed-Form Bid Approximation}

This problem is a variant of the multi-vehicle asymmetric routing problem with energy objectives, which is NP-hard~\cite{toth2002vrp}. We solve it via a two-stage hierarchical decomposition.

For candidate task~$t$ and robot~$i$ with current end position~$q_i$, the bid consists of unloaded transit energy (payload $w=0$) plus loaded task-execution energy (payload $w = w_t$):
\begin{equation}
b_i(t) = E_{\mathrm{approx}}(q_i, p_t, 0) + E_{\mathrm{approx}}(p_t, d_t, w_t)
\label{eq:bid}
\end{equation}
where $E_{\mathrm{approx}}$ is the closed-form energy approximation from~\eqref{eq:eseg}:
\begin{equation}
\begin{aligned}
E_{\mathrm{approx}}(a,b,w) = \frac{1}{\eta}\Big[&\mu(M\!+\!w)g\,\|b-a\| \\
&+ \tfrac{1}{2}(M\!+\!w)\bigl(v_f^2 - v_0^2\bigr)^{\!+}\Big]
\end{aligned}
\label{eq:eapprox}
\end{equation}
where $(x)^+ = \max(0, x)$. Note that unloaded transit uses $w = 0$ while loaded execution uses $w = w_t$, the task payload mass. Each bid computation is $O(1)$.

\section{Sequential Auction Allocation}

The allocation stage uses a centralized sequential single-item auction~\cite{koenig2006power}. In each round, one unassigned task is awarded to the robot with the lowest energy bid. Algorithm~\ref{alg:auction} details the procedure.

\begin{algorithm}[t]
\caption{Energy-Aware Sequential Auction}
\label{alg:auction}
\begin{algorithmic}[1]
\REQUIRE Robots $\mathcal{N}$, tasks $\mathcal{T}$, energy params $\theta$
\ENSURE Assignment $\{\sigma_i\}_{i \in \mathcal{N}}$
\STATE Initialize $\sigma_i \leftarrow ()$, $q_i \leftarrow \mathrm{depot}_i$ for all $i$
\STATE $\mathcal{U} \leftarrow \mathcal{T}$
\WHILE{$\mathcal{U} \neq \emptyset$}
    \STATE $b^* \leftarrow \infty$, $i^* \leftarrow \mathrm{null}$, $t^* \leftarrow \mathrm{null}$
    \FOR{each task $t \in \mathcal{U}$}
        \FOR{each robot $i \in \mathcal{N}$}
            \STATE $b_i(t) \leftarrow E_{\mathrm{approx}}(q_i, p_t, 0; \theta) + E_{\mathrm{approx}}(p_t, d_t, w_t; \theta)$
            \IF{$b_i(t) < b^*$}
                \STATE $b^* \leftarrow b_i(t)$, $i^* \leftarrow i$, $t^* \leftarrow t$
            \ENDIF
        \ENDFOR
    \ENDFOR
    \STATE Append $t^*$ to $\sigma_{i^*}$
    \STATE $q_{i^*} \leftarrow d_{t^*}$
    \STATE $\mathcal{U} \leftarrow \mathcal{U} \setminus \{t^*\}$
\ENDWHILE
\RETURN $\{\sigma_i\}_{i \in \mathcal{N}}$
\end{algorithmic}
\end{algorithm}

The auction runs $m$ rounds, each evaluating $n$ bids across the remaining tasks, yielding $O(n \cdot m^2)$ total complexity.

By construction, each bid $b_i(t)$ equals the marginal cost of a feasible appended schedule under the closed-form model~\eqref{eq:eapprox}, so the winning robot is never assigned a task whose declared cost exceeds that of the feasible continuation used to compute the bid. The auction is also monotone: lowering a winning bid while holding all competing bids fixed cannot reverse the outcome. If the per-bid approximation error is bounded by $\bar{\varepsilon}$, the auction's total fleet cost is within $2m\bar{\varepsilon}$ of any feasible sequential allocation. In practice this worst-case additive bound is loose---the empirical mean per-bid error is 15.7\%, yielding a relative bound of roughly 82\% for $m=50$ tasks---but the tighter 4.5\% worst-case empirical gap observed at small scale indicates that bid errors partially cancel across rounds.

\section{Per-Robot Trajectory Generation and Event-Triggered Rescheduling}

\subsection{Stage 2: Per-Robot Trajectory Generation}

After allocation, each robot independently solves the optimal control problem (OCP) from~\cite{bakshi2021energy} for its assigned task sequence. This second stage is precisely the continuous optimization problem in \eqref{eq:single_ocp_main}--\eqref{eq:single_ocp_time}, or equivalently the reduced statement in \eqref{eq:ocp} when only the trajectory generation stage is emphasized. The output is the true energy-optimal state-control trajectory for the allocated sequence.

\subsection{Hybrid Trigger Logic}

Three event types trigger partial rescheduling. Let $t_k$ denote the $k$th rescheduling instant and let $t_k^+$ denote the state immediately after the corresponding reallocation is committed. For each robot~$i$ and each newly arriving priority task~$t^{\mathrm{p}}$, define the indicator signals
\begin{align}
\mathbf{1}^{\mathrm{fault}}_i(t) &= \begin{cases}
1, & \text{if robot $i$ becomes unavailable at time } t\\
0, & \text{otherwise}
\end{cases} \label{eq:fault_indicator}\\
\mathbf{1}^{\mathrm{prio}}_{t^{\mathrm{p}}}(t) &= \begin{cases}
1, & \text{if priority task $t^{\mathrm{p}}$ arrives at time } t\\
0, & \text{otherwise}
\end{cases} \label{eq:prio_indicator}\\
\mathbf{1}^{\mathrm{dev}}_i(t) &= \begin{cases}
1, & \text{if } \dfrac{|E^{\mathrm{act}}_i(t)-E^{\mathrm{pred}}_i(t)|}{E^{\mathrm{pred}}_i(t)} > \delta \\[4pt]
   & \text{and } t-t_{k^\ast(i)} \ge \Delta t_{\min}, \\[6pt]
0, & \text{otherwise.}
\end{cases} \label{eq:dev_indicator}
\end{align}
where $k^\ast(i)$ is the index of the most recent energy-deviation reschedule involving robot~$i$. The global switching signal is
\begin{equation}
\Gamma(t)=
\mathbf{1}\!\left\{
\max_i \mathbf{1}^{\mathrm{fault}}_i(t)
\vee
\max_{t^{\mathrm{p}}}\mathbf{1}^{\mathrm{prio}}_{t^{\mathrm{p}}}(t)
\vee
\max_i \mathbf{1}^{\mathrm{dev}}_i(t)
=1
\right\}.
\label{eq:global_trigger}
\end{equation}
Whenever $\Gamma(t)=1$, only the unstarted tasks of the affected robots, together with any newly inserted priority tasks, are re-auctioned using current robot states as the new initial conditions.

Table~\ref{tab:events} summarizes the trigger semantics without changing the hybrid law in \eqref{eq:fault_indicator}--\eqref{eq:global_trigger}.

\begin{table}[t]
\centering
\caption{Rescheduling Event Types}
\label{tab:events}
\begin{tabular}{lll}
\toprule
\textbf{Event} & \textbf{Trigger} & \textbf{Action} \\
\midrule
Robot fault & Robot $i$ unavailable & Re-auction $\sigma_i$ unstarted \\
Priority task & New task $t_{\mathrm{new}}$ & Auction $t_{\mathrm{new}}$ \\
Energy dev. & $|E_{\mathrm{act}} - E_{\mathrm{pred}}|/E_{\mathrm{pred}} > \delta$ & Re-auction deviant robot \\
\bottomrule
\end{tabular}
\end{table}

The trigger thresholds are fixed across all 505 scenarios: $\delta = 0.10$ (10\% energy deviation) and $\Delta t_{\min} = 5$\,s (minimum inter-event interval for energy-deviation triggers). These values were selected once via preliminary sensitivity analysis on a held-out configuration set and held constant throughout all experiments; no per-scenario tuning was performed. The minimum inter-event interval $\Delta t_{\min}$ in~\eqref{eq:dev_indicator} ensures that the total number of rescheduling events on any finite horizon $[0,T]$ is bounded: each fault or priority-task arrival triggers at most one reschedule, and each robot can produce at most $\lfloor T/\Delta t_{\min}\rfloor+1$ energy-deviation triggers. Thus infinitely many rescheduling events cannot accumulate in finite time (i.e., Zeno behavior is excluded by design).

\subsection{Warm-Start Feasibility and Practical Cost-to-Go Bound}

Let $\mathcal{U}_k$ denote the set of unstarted tasks immediately before the $k$th rescheduling event, and let $\Sigma_k^{-}$ be the pre-event schedule truncated to those unstarted tasks. Let $\Xi_k$ collect the robot states at the trigger time. Define the predicted remaining cost-to-go
\begin{equation}
V_k := \sum_{i=1}^n \widehat{J}_{i,\mathrm{rem}}(\Sigma_k,\Xi_k) + \rho |\mathcal{U}_k|,
\label{eq:Vk}
\end{equation}
where $\widehat{J}_{i,\mathrm{rem}}$ is the warm-start prediction of remaining trajectory energy and $\rho\ge 0$ is a bookkeeping weight for pending tasks. The warm start is the feasible schedule obtained by retaining completed task prefixes and using $\Sigma_k^{-}$ as the initial guess for the local trajectory solver on the remaining suffix.

As long as at least one robot remains capable of executing each remaining task, the truncated pre-event schedule $\Sigma_k^{-}$ is feasible for the post-event problem, so the warm-started rescheduling problem is always initialized from a feasible point. Furthermore, assuming the warm-start trajectory solver is descent preserving (i.e., the post-reschedule plan has predicted remaining cost no greater than its initialization) and that $\widehat{J}_{i,\mathrm{rem}}$ is locally Lipschitz in $\Xi_k$ with constant $L_J$, the cost-to-go sequence satisfies the practical descent inequality
\begin{equation}
V_{k+1} \le V_k + L_J \|\Xi_{k+1}-\Xi_k^+\| - \rho\,\Delta N_k,
\label{eq:lyap_bound}
\end{equation}
where $\Delta N_k$ is the number of tasks completed between consecutive rescheduling instants and $\Xi_k^+$ is the immediately post-reschedule state. This is a \emph{practical descent condition}, not a formal closed-loop stability proof. It provides a sufficient condition under which warm-start rescheduling preserves feasibility and prevents uncontrolled growth of the predicted remaining cost. It does not constitute a stability guaranty or a worst-case performance bound for the overall fleet trajectory.

\section{Simulation Results}

We implemented the framework in Python~3.12 on an Intel i7-12700H processor (single-threaded) and evaluated it across 505 simulation runs (101 configurations $\times$ 5 random seeds). The workspace is a $20\,\mathrm{m} \times 20\,\mathrm{m}$ factory floor with three layout types: grid (regular stations), random (uniformly distributed), and clustered (3--5 manufacturing cells). Robot parameters follow the nominal values in~\cite{bakshi2021energy}: $M = 50$\,kg, $v_{\max} = 1.5$\,m/s, $\eta = 0.85$, $\mu = 0.02$, $w_{\max} = 20$\,kg. Results are conditioned on these calibration assumptions; hardware transfer requires platform-specific parameter identification. We compare against four baselines of increasing strength: B1~(nearest-task heuristic with OCP trajectories~\cite{bakshi2021energy})---a lightweight operational heuristic commonly used in practice; B2~(nearest-robot with constant-velocity paths)---isolating the value of trajectory optimization; B3~(same sequential auction with Euclidean distance bids)---the primary diagnostic comparison for bid-metric selection; and B4~(exhaustive $n^m$ assignment enumeration at small scale, $n \leq 3$, $m \leq 8$, fixed task ordering).

\begin{figure}[t]
\centering
\includegraphics[width=\columnwidth]{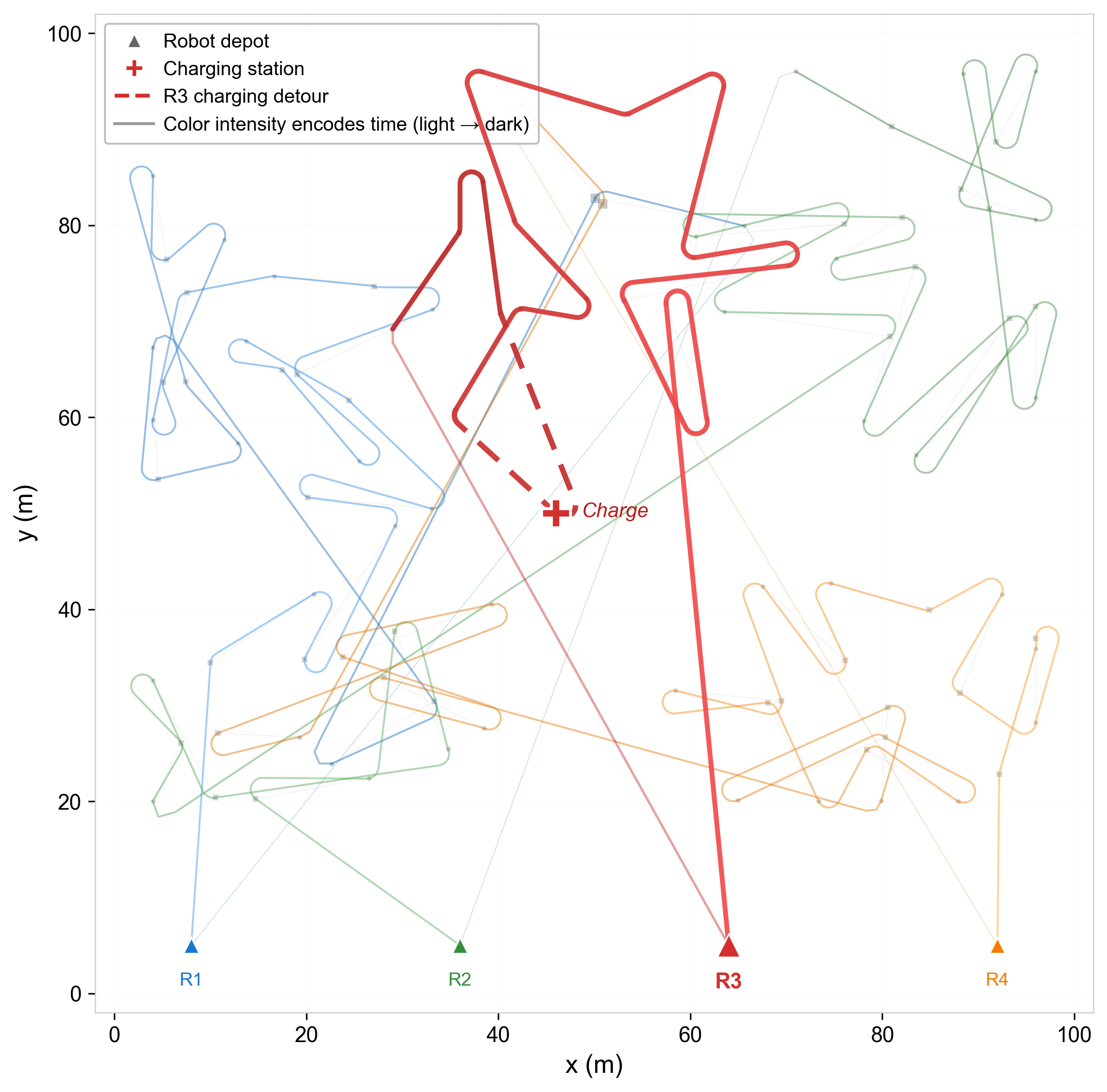}
\caption{Fleet trajectory for 4~AMRs and 50~tasks on a clustered layout.}
\label{fig:trajectory}
\end{figure}

Fig.~\ref{fig:trajectory} illustrates a representative scenario with 4~AMRs and 50~tasks on a clustered layout. Robot~R3 undergoes a mid-route charging event, triggering the rescheduling of its remaining tasks to R1 and R4.

\textbf{Fleet energy savings.} Table~\ref{tab:savings} reports energy savings across fleet sizes (proposed vs.\ B1 significant at $p<0.01$ for $n \geq 5$; proposed vs.\ B3 significant deficit at $p<0.05$ for all sizes, Wilcoxon signed-rank). Both auction variants achieve 11.8\% average savings over the nearest-task heuristic~(B1), scaling from 2.7\% at $n=5$ to 25.4\% at $n=20$ as larger fleets offer more reallocation flexibility. The B1 comparison demonstrates the value of auction-based allocation over a simple operational heuristic. Against B2~(nearest-robot, constant-velocity), savings are 9--18\%, isolating the additional benefit of OCP trajectory optimization.

\begin{table}[t]
\centering
\caption{Fleet Energy Savings (\%)}
\label{tab:savings}
\begin{tabular}{lcccc}
\toprule
Fleet & Energy & vs.\ B1 & vs.\ B2 & vs.\ B3 \\
Size & (kJ) & Nearest & Short-path & Dist.\ Auc. \\
\midrule
2  & 1.18 & $-3.3 \pm 1.8$ & $+17.9 \pm 2.4$ & $-4.1 \pm 1.2$ \\
5  & 2.45 & $+2.7 \pm 2.1$ & $+10.3 \pm 1.9$ & $-4.0 \pm 1.4$ \\
10 & 3.82 & $+10.3 \pm 2.6$ & $+9.2 \pm 2.2$  & $-3.9 \pm 1.1$ \\
15 & 4.96 & $+23.7 \pm 3.1$ & $+11.6 \pm 2.5$ & $-2.7 \pm 0.9$ \\
20 & 5.71 & $+25.4 \pm 3.4$ & $+9.2 \pm 2.0$  & $-2.6 \pm 0.8$ \\
\midrule
Avg. & --- & $+11.8$ & $+11.6$ & $-3.5$ \\
\bottomrule
\end{tabular}
\end{table}

\textbf{Energy vs.\ distance bids.} In the uniform-friction workspaces tested, the energy-based method performs 2.6--4.1\% worse than B3. The closed-form approximation~\eqref{eq:eapprox} introduces 15.7\% mean error relative to the full OCP, yielding 87\% winner identification accuracy (91\% grid, 86\% random, 84\% clustered). Ranking errors concentrate on short segments ($s < 2$\,m) and near-tied bids ($|b_i - b_j| < 5\%$). The causal mechanism is: in uniform-friction workspaces, energy is nearly proportional to distance ($r > 0.95$), so distance captures the correct bid ranking with zero approximation error, while the closed-form surrogate~\eqref{eq:eapprox} introduces rank-order inversions in 13\% of auction rounds that accumulate across rounds.

\textbf{Friction heterogeneity crossover.} Table~\ref{tab:heterogeneity} isolates the effect of workspace friction variation on bid-metric selection. In uniform terrain ($\mu=0.02$, $r=0.98$), distance bids outperform energy bids by 3.7\%, as the closed-form surrogate's approximation error degrades ranking accuracy while distance captures the correct ordering with no model error. As friction heterogeneity increases, the energy--distance correlation~$r$ drops and distance loses its ranking fidelity: at $\mu\in[0.01,0.04]$ ($r=0.84$) the two metrics are statistically tied, and for $\mu\in[0.005,0.06]$ ($r=0.75$) the zone-aware energy bid yields 2.1\% savings with 91\% bid accuracy. The widest friction range ($\mu\in[0.005,0.08]$, $r=0.74$) extends the energy-bid advantage to 2.4\% at 92\% accuracy. The crossover occurs near $r\approx 0.85$, providing a practical decision rule: practitioners should measure or estimate the energy--distance correlation for their workspace and select the bid metric accordingly. The central result is that both auction variants substantially outperform B1 and B2, demonstrating the value of the auction-plus-OCP structure itself, while the choice between energy and distance bids is a regime-dependent design decision.

\begin{table}[t]
\centering
\caption{Energy vs.\ Distance Bids Under Friction Heterogeneity}
\label{tab:heterogeneity}
\begin{tabular}{lcccc}
\toprule
Friction Range & $r$ & Gap (\%) & Bid Acc. & Winner \\
\midrule
Uniform ($\mu=0.02$) & $0.98$ & $+3.7 \pm 3.2$ & 89\% & Distance \\
$\mu \in [0.015, 0.03]$ & $0.93$ & $+2.5 \pm 2.9$ & 87\% & Distance \\
$\mu \in [0.01, 0.04]$ & $0.84$ & $+0.7 \pm 4.4$ & 88\% & Tie \\
$\mu \in [0.005, 0.06]$ & $0.75$ & $-2.1 \pm 6.3$ & 91\% & Energy \\
$\mu \in [0.005, 0.08]$ & $0.74$ & $-2.4 \pm 6.6$ & 92\% & Energy \\
\bottomrule
\end{tabular}
\end{table}

\textbf{Computation and rescheduling.} Table~\ref{tab:computation} breaks down the pipeline: auction allocation scales as $O(nm^2)$, while per-robot OCP dominates total time (e.g., 1.2\,s of 1.7\,s total for $m=100$). Rescheduling latency (Table~\ref{tab:rescheduling}) averages 5.6\,ms with only 6\% energy overhead versus cold-start full re-auction, offering a favorable speed–quality tradeoff for real-time operation.

\begin{table}[t]
\centering
\caption{Computation Time Breakdown ($n=10$)}
\label{tab:computation}
\begin{tabular}{lccc}
\toprule
Tasks ($m$) & Auction (ms) & OCP (ms) & Total (ms) \\
\midrule
10  & 1.7  & 48.3   & 52.1 \\
20  & 6.3  & 124.7  & 136.8 \\
50  & 43.3 & 387.2  & 448.9 \\
100 & 405.2 & 1241.6 & 1699.1 \\
\bottomrule
\end{tabular}
\end{table}

\begin{table}[t]
\centering
\caption{Rescheduling Performance by Disruption Type}
\label{tab:rescheduling}
\begin{tabular}{lcccc}
\toprule
Disruption & Mean & Max & Tasks & Overhead \\
Type & (ms) & (ms) & Reassgn. & vs.\ Cold (\%) \\
\midrule
Robot fault       & 4.2 & 7.8 & 6.3 & $+8.2$ \\
Priority task     & 7.1 & 9.0 & 1.0 & $+3.1$ \\
Energy deviation  & 3.5 & 6.2 & 3.8 & $+5.7$ \\
Combined          & 6.7 & 7.3 & 5.1 & $+6.9$ \\
\midrule
Overall           & 5.6 & 9.0 & 4.1 & $+6.0$ \\
\bottomrule
\end{tabular}
\end{table}

\textbf{Gap to enumerated assignment benchmark.} At a small scale ($n \leq 3$, $m \leq 8$), the auction achieves a worst-case gap of 4.5\% (average 2.1\%) relative to B4, which exhaustively enumerates all $n^m$ robot-to-task assignments but uses a fixed task ordering within each robot. Because B4 does not search over task orderings, its cost is an upper bound on the global combinatorial optimum; consequently, the measured gaps are a lower bound on the true optimality gap over both assignment and sequencing.

\section{Conclusion}

We presented a hierarchical framework coupling auction-based task allocation with per-robot energy-conscious trajectory optimization for AMR fleets in asymmetric task spaces. Both auction variants achieve 11.8\% average energy savings over nearest-task allocation (up to 25.4\% at 20~robots), with event-triggered rescheduling under 10\,ms and a 4.5\% worst-case gap to the enumerated benchmark.

The central finding is regime-dependent bid-metric performance. In uniform terrain, distance bids outperform energy bids by 3.5\% ($p < 0.05$, Wilcoxon) because the 15.7\% closed-form approximation error inverts bid rankings in 13\% of rounds while energy and distance remain highly correlated ($r > 0.95$). When friction heterogeneity is sufficient ($r < 0.85$), a zone-aware energy bid recovers 2--2.4\% savings with ranking accuracy rising to 92\%, as detailed in Table~\ref{tab:heterogeneity}. Practitioner guidance: use distance bids in near-uniform terrain and energy-aware bids when friction variation is significant. Future work will address hardware validation and collision-aware allocation.

\section*{Acknowledgment}
Claude was used to assist with the language editing of this manuscript.

\bibliographystyle{IEEEtran}
\bibliography{references}

\end{document}